# 集成领域知识的"弱监督强特征"-"强监督弱特征"级联分类方法遥感作物类型制图实践

臧运泽[1] 刘亦飞[1] 陈学泓[1] 李安琪[1] 翟依晨[1] 李世捷[1] 刘璐玲[1] 朱传海[1] 陈瑞麟[1] 李淑鹏[1] 揭娜[1]

1. 北京师范大学 地理学部 遥感科学与工程研究院，北京 100875

## 1.问题背景

利用卫星遥感技术对农作物进行识别和监测，对优化生产布局、调整生产方式具有重要意义。近年来，随着越来越多的遥感卫星发射运行以及更多的遥感数据开源，遥感作物制图可用的数据资源逐渐丰富而成本逐步变低（Huang et al，2018）；同时，近年来机器学习与人工智能领域快速发展，涌现了诸多性能优异的分类模型或算法，如随机森林、深度神经网络等，已广泛用于的作物分类（Zhong et al, 2019；吴立周等，2020；Adrian，2021）。以上两方面发展为遥感作物识别提供了优异的数据与技术条件（陈仲新等，2016）。然而，其中的大多数机器学习方法十分依赖于充足的样本（Kamilaris，2018）；而获得足量、准确的样本往往需要展开大量的目视识别、标记、及实地核验工作，十分费时费力，同时不同区域不同时间的作物特征差异又进一步加剧了样本需求与获取代价。因此，在小样本/无样本情形下，则需要领域知识或专家知识提供重要信息来源。本研究在团队之前提出的"弱监督强特征"-"强监督弱特征"级联分类框架下，进一步优化领域知识集成过程，实现无样本条件下的作物分类。

"弱监督强特征"-"强监督弱特征"级联分类框架是团队此前设计的一种面向小样本/无样本的作物制图框架，其主要思想是针对不同地区不同类型遥感数据可获取性存在差异的特点，混用强弱监督分类器达到样本扩增的目的。一般来说，农作物反射光谱的季节性变化（即时序特征）与其在高分辨率影像中的纹理（即空间特征）是用以区分作物种类的有效特征。当具备完整清晰的高时空分辨率影像时，作物间差异显著的分类特征通常也容易被确定，因而在该情形下容易根据少量典型样本或领域专家知识构建足够准确的分类器(Qiu et al,2017; Pan et al, 2012; Belgiu et al，2018)。我们称这种需要少量特征显著的样本或知识参与分类的情形为"弱监督强特征"情形。然而，对于大范围作物制图而言，由于云遮挡现象十分普遍且部分观测的质量欠佳，一般采用的以哨兵系列为代表的中分辨率遥感数据存在相当程度的数据缺失与光谱失真，以及部分地区的高分辨率影像数据不足，这导致用以区分作物的时空特征被弱化或复杂化，因而通常需要大量的样本参与训练才能构建有效的分类器（Xu et al，2020；You et al，2021）。我们称这种需要大量特征不显著样本参与分类的情形为"强监督弱特征"情形。由于不同地区遥感数据的缺失程度及高分影像的可获取性存在差异，"弱

监督强特征"-"强监督弱特征"级联分类框架首先利用专家知识将小部分时空完整性相对较好、且存在"强特征"的像元进行弱监督分类，从而对于这小部分"强特征"像元得到优质的分类结果。接着，我们将这些"强特征"像元通过特征弱化的方式自动生成大量"弱特征"样本输入强监督分类器，最终实现在少样本或无样本情况下的作物分类。

在"弱监督强特征"-"强监督弱特征"级联分类框架下，本研究进一步优化领域知识的集成过程，主要在以下方面进行改进：在特征弱化步骤，充分考虑云分布情况，生成更符合实际数据缺失分布的弱特征样本。

# 2.技术解决方案

## 2.1 "弱监督强特征"-"强监督弱特征"级联分类框架原理

用于大范围作物制图的遥感数据存在相当程度的数据缺失，而且不同地区数据的缺失程度也不一致，如图 1 中遥感影像所示，部分像元的时间序列较完整（如像元 A、E、G），部分像元时间序列上的无效观测较多（如像元 B、C、D、H）。当某像元时间序列较完整时，我们可以从中得到该像元对应作物完整的时序特征，如(a)右侧曲线 A。如果我们能通过专家知识确定该作物的类型，我们便称这样的时间序列为"强特征"样本。反之，当某像元时间序列上观测残缺较多时，我们很难获得其时序特征，如（a）右侧曲线 C，我们称这样的时间序列为"弱特征"样本。

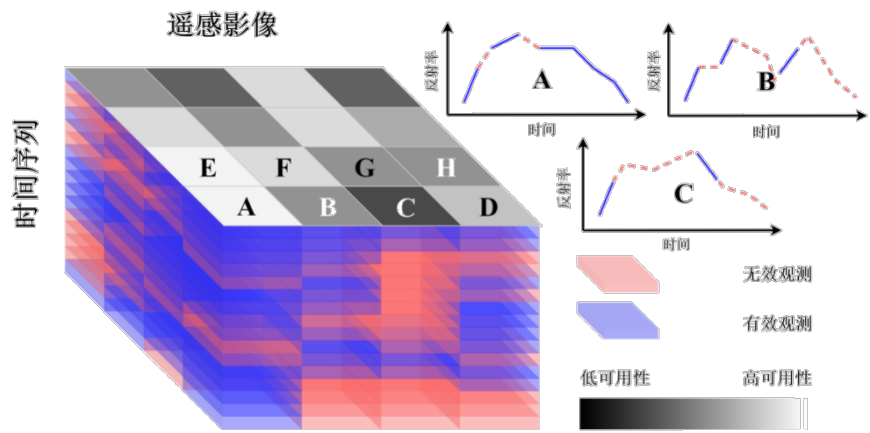

图 1 "强特征"和"弱特征"样本

一般来说，如若我们拥有数量充足且标记准确的训练样本，现存的很多作物制图算法可完成分类任务。但在实际中，标记足量的样本往往需要展开大量的调查、目视识别、及实地核验工作，十分耗时费力，其技术经济性、场景通用性均不高。因此，本研究提出了此级联算法框架，其优势在于可在无样本的情形下基于专家知识决策、特征弱化算法生成大量可运用于强监督分类器的样本。其原理如图 2 所示：

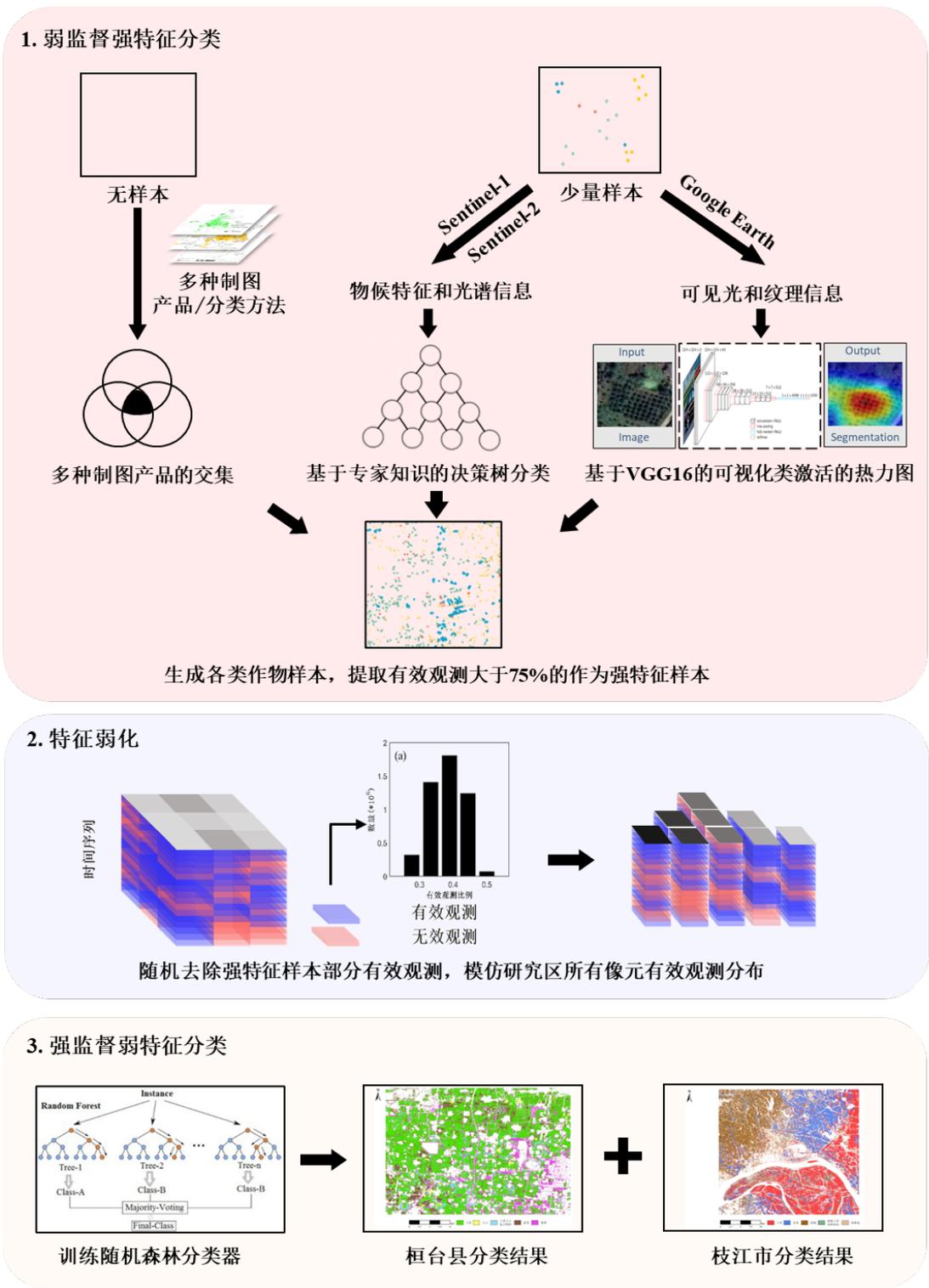

图 2 "弱监督强特征"-"强监督弱特征"级联分类框架

当我们拥有"强特征"时间序列时，只需使用弱监督分类器便可为其标记足够准确的标签，从而获得"强特征"样本。首先根据物候信息、光谱特征以及高分影像上的纹理信息，由文献调研和专家经验经过决策树和深度学习网络的筛选，生成各类作物样本，保留了拥有75%以上有效观测的时序作为待标记（或待分类）的"强特征"样本集合。但对于大范围的作物制图，单纯利用基于专家知识的弱监督分类器难以解决问题。这是由于"弱特征"时序

数量占大多数，而弱监督分类器只能胜任对"强特征"时序的分类任务，对于大多数由于缺少部分有效观测的"弱特征"时序，由于样本太少且分类特征不明显等原因难以获得理想的结果。因此，本研究还需依靠强监督分类器完成最后的制图工作，而这就意味着我们需要大量"弱特征"样本训练某种强监督分类器。在样本生成的过程中，随机剔除了部分"强特征"样本时序中的有效观测，使其弱化为"弱特征"样本以模仿研究区真实情况，参与强监督分类器的训练。最后，本研究采用随机森林作为强监督分类器进行作物分类制图。

## 2.2 集成领域知识的"弱监督强特征"-"强监督弱特征"级联分类方法改进

在特征弱化步骤，充分考虑云分布情况，生成更符合实际数据缺失分布的弱特征样本。研究区所有像元的有效观测分布代表了云分布情况，当研究区处于常年云覆盖的情况下，有效观测少的像元占主导，而有效观测多的像元占少数。而研究区常年无云时，有效观测多的像元占主导。因此若不对"强特征"样本处理，或者仅仅按照固定的比例剔除"强特征"样本时序中的有效观测，难以模拟出研究区的真实情况。因此，本研究统计了研究区内所有像元的有效观测数量分布直方图，按照像元数量比例和对应的有效观测数，随机去除某一数量比例"强特征"样本时序中的某一观测数量的有效观测，以生成符合实际数据缺失的"弱特征"样本。

## 2.3 研究区分类实践

### 2.3.1 数据及预处理

枝江研究区使用了 2022 年 04 月 15 日-2022 年 09 月 20 日的 Sentinel-1 和 Sentinel-2 的影像数据，桓台研究区使用了 2022 年 06 月 01 日-2022 年 09 月 20 日的 Sentinel-1 和 Sentinel-2 的影像数据。影像数据详细信息见表 1。其中，我们将 Sentinel-1 数据合成为以 12 天为间隔的时间序列。选取了云量小于 95% 的 Sentinel-2 影像并将其合成为以 10 天为间隔的时间序列，并采用线性插值填补了时间序列上的空缺值。

表 1 哨兵数据简介

| | 波段 | 中心波长（$\mu m$） | 分辨率（m） |
| --- | --- | --- | --- |
| | Band 2 - Blue | 0.49 | 10 |
| | Band 3 - Green | 0.56 | 10 |
| | Band 4 - Red | 0.665 | 10 |
| **Sentinel-2** | Band 5 – Red Edge 1 | 0.705 | 20 |
| | Band 6 - Red Edge 2 | 0.74 | 20 |
| | Band 7 - Red Edge 3 | 0.783 | 20 |
| | Band 8 - NIR | 0.842 | 10 |

| | Band 8A -Red Edge 4 | 0.865 | 20 |
|---|---|---|---|
| | Band 11 - SWIR | 1.61 | 20 |
| | Band 12 – SWIR2 | 2.19 | 20 |
| Sentinel-1 | VV | / | 10 |
| | VH | / | 10 |

注：数据来源于 Google Earth Engine 平台，Sentinel-1 为"COPERNICUS/S1_GRD"产品， Sentinel-2 为"COPERNICUS/S2_SR_HARMONIZED"产品。

此外，基于上述经预处理的数据，本研究逐时相计算了每幅影像的各类指数（NDVI、EVI、LSWI、MNDWI、NDYI）用于强弱监督分类器分类。各指数的计算公式参见表 2。

表 2 计算指数简介

| 指数 | 表达式 |
|---|---|
| NDVI | $NDVI = \dfrac{\rho_{nir} - \rho_{red}}{\rho_{nir} + \rho_{red}}$ |
| EVI | $EVI = 2.5 * \dfrac{\rho_{nir} - \rho_{red}}{\rho_{nir} + 6 * \rho_{red} - 7.5 * \rho_{blue} + 1}$ |
| LSWI | $LSWI = \dfrac{\rho_{nir} - \rho_{swir}}{\rho_{nir} + \rho_{swir}}$ |
| MNDWI | $MNDWI = \dfrac{\rho_{green} - \rho_{swir}}{\rho_{green} + \rho_{swir}}$ |
| NDYI | $NDYI = \dfrac{\rho_{green} - \rho_{blue}}{\rho_{green} + \rho_{blue}}$ |

**2.3.2 弱监督强特征分类**

级联算法框架的第一部分需要借助弱监督分类器标记一些时序上较完整的作物样本以备后用。因此，我们首先筛选出了研究区内若干个有效观测数大于作物识别区域内所有像元有效观测数 75%分位数的像元作为待标记的"强特征"像元集合。我们依据数据预处理时计算得到的各类指数，并利用基于专家知识的决策树分类器对该集合中的"强特征"时序进行标记，得到"强特征"样本集合。上述过程如图 2 所示。决策树的设定规则主要依据极少量现地调查与已有文献分析，详细规则说明请参见附录。需要说明的是，由于该分类结果将作为第二阶段"强监督弱特征"分类的样本，因此其规则阈值设定偏向严格以尽可能减少错分但容许漏分，

这种阈值设定原则在满足本方法需求的同时减少了对阈值准确性的依赖。

### 2.3.3 特征弱化及样本生成策略

根据"弱监督强特征"分类的结果，如图 3 和图 4 所示，本研究在比赛官方给定的研究区周围生成 30 公里缓冲区，在桓台县所在的缓冲区中选取了 2000 个玉米样本，2000 个大豆样本，2000 个芹菜样本，2000 个玉米大豆混种样本和 1000 个果树样本。此外，依据 Global 30 中的各种非耕地类别（林地、草地、湿地、水体和不透水层）进行分层抽样，选取 1000 个像元作为其它类别样本。提取各作物样本及其它类别样本在作物生长季节的 Sentinel-1 和 Sentinel-2 的各波段的时间序列作为分类的特征，我们将这一步骤获取的样本称为"强特征"样本。

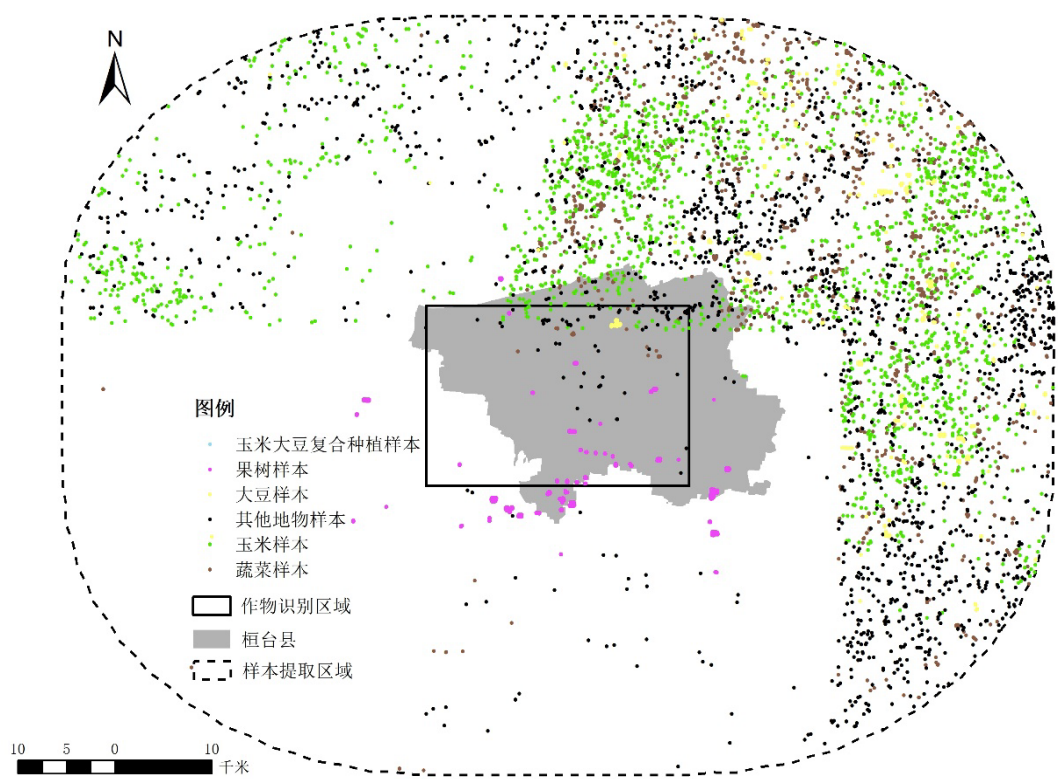

图 3 在桓台县周围 30 公里缓冲区筛选的"强特征"样本

同样，在枝江市所在的缓冲区中选取了 2000 个玉米样本，2000 个水稻样本，2000 个休耕地样本，2000 个玉米柑橘混种样本和 1000 个柑橘样本。依据 Global 30 中的各种非耕地类别（林地、草地、湿地、水体和不透水层）进行分层抽样，选取 1000 个像元作为其它类别样本。

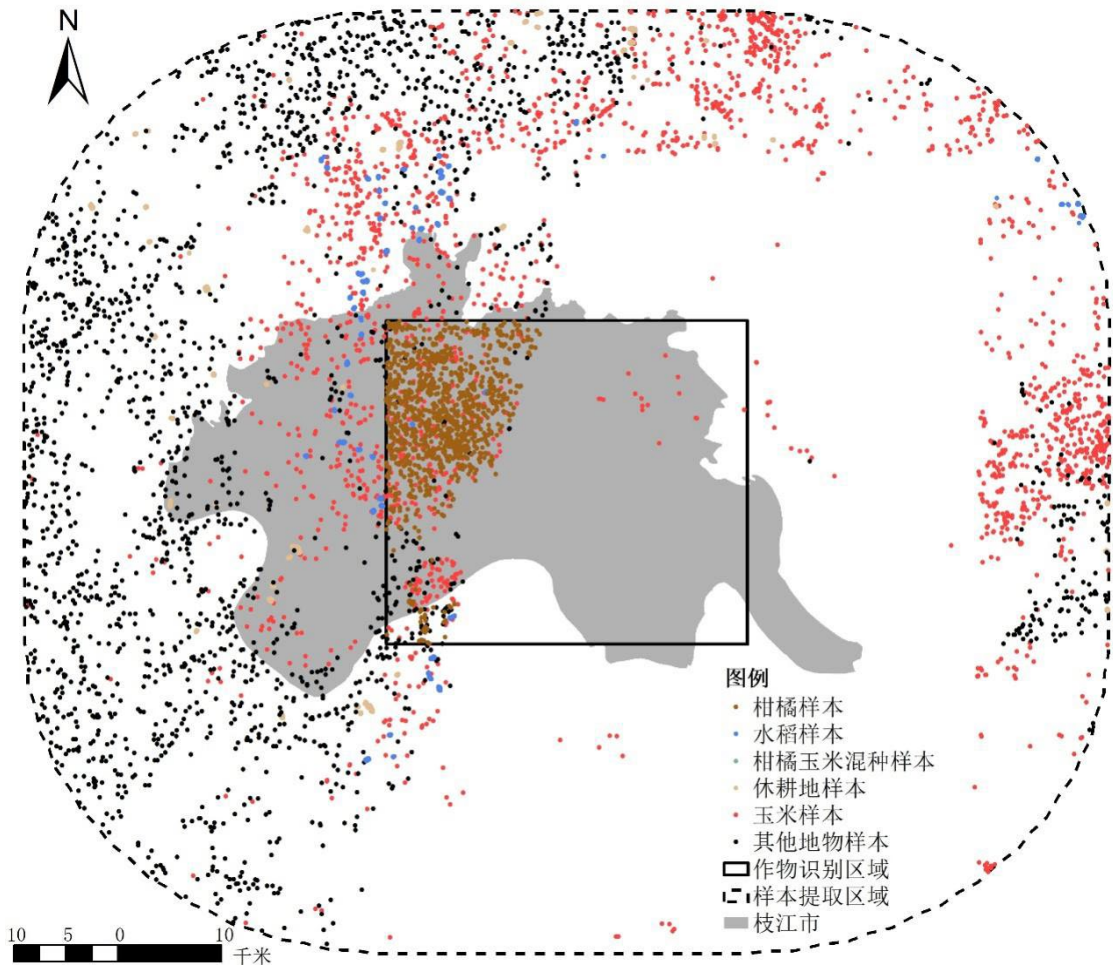

图 4 在枝江市周围 30 公里缓冲区筛选的"强特征"样本

由于选取的样本均具有有效观测数高和明显的特征，而分类区域中的像元质量较低。如果直接使用高质量样本进行训练，则分类器的泛化性能欠缺，容易造成样本代表性不足。因此，需要将高质量的样本进行特征弱化，模拟分类区域中平均质量水平的样本。本研究采用了降低有效观测数的方法处理待识别作物的高质量样本。

如图 5 所示，选取样本时间序列的有效观测数均较高(图 5(a))，而研究区域内像元的有效观测比例参差不齐(图 6)。为了增加分类器的泛化性，本研究随机去除高质量样本的有效观测，使样本有效观测比例分布与研究区内各作物生长季内有效观测比例分布相同。

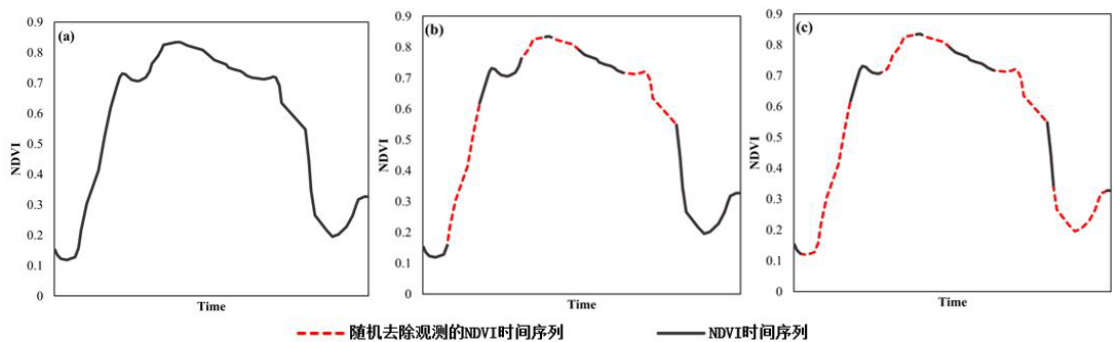

图 5 随机去除有效观测示意图。(a)原始 NDVI 时间序列；(b)随机去除 25%的有效观测; (c)随机去除 50%的有效观测

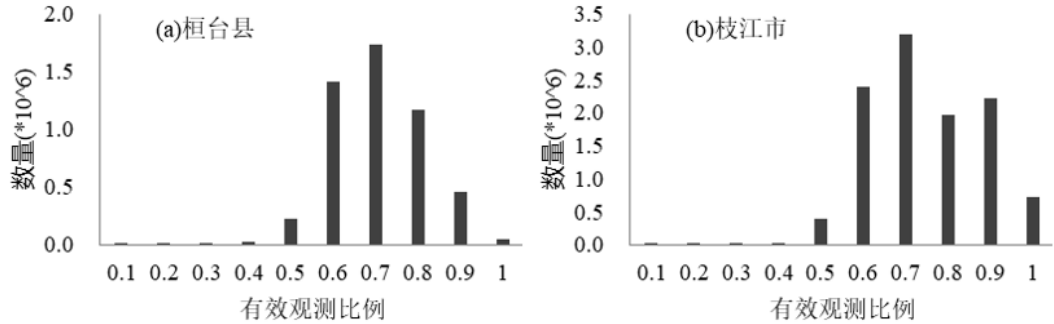

图 6 作物识别区域内像元有效观测比例分布直方图。(a) 山东桓台作物识别区域内 2022 年 06 月 01 日-2022 年 09 月 20 日有效观测比例分布直方图；(b) 湖北枝江作物识别区域内 2022 年 04 月 15 日- 2022 年 09 月 20 日有效观测比例分布直方图

经过样本弱化后，在湖北枝江得到玉米、水稻、休耕地、柑橘玉米混种和其它类样本各 4000 个。

在山东桓台得到大豆、玉米、大豆玉米混种、蔬菜和其它类样本各 4000 个，其中其它类别样本由果树、其他样本等比例混合而成。考虑到混种作物的实际面积，随机筛选了 500 个混种样本用于分类。另外每个类别加入了少量真实样本（30 个/类）作为下一步"强监督弱特征分类"的训练样本。

### 2.3.4 强监督弱特征分类

基于 2.2.3 节获取的大量样本，本研究使用随机森林分类器进行作物分类。本研究使用湖北枝江作物样本集用于玉米、水稻、玉米柑橘混种、休耕地的分类，使用山东桓台作物样本集用于大豆、蔬菜、玉米、玉米大豆混种的分类。为了确定分类器的超参数——随机森林的树个数，样本集都以 7：3 的比例划分为训练集和验证集。依次确定分类器的最优超参数，如图 7 所示，随着树的个数不断增加，分类器在验证集上的精度先增加后趋于稳定。分类器在验证集上最高精度所对应的树个数确定为最终随机森林的参数。如图 7 所示，山东桓台作物分类器的树个数为 110，湖北枝江作物分类器的树个数为 290。

考虑到柑橘和果树的负类包含多个树种，获取负样本较为困难，因此柑橘和果树的分类使用单类分类器分类。本研究使用 PUL（Positive and unlabeled learning）策略训练随机森林分类器。 PUL 将传统的二分分类问题转化为了单分类问题，通过在研究区域随机选取未标记样本并估计二类分类概率的策略，克服了负样本难以获取的问题。本研究使用柑橘样本集和果树样本集分别训练单类分类器用于柑橘和果树的分类。由强特征生成的果树/柑橘样本作为标记样本，同时在研究区中随机采样相同数量的未标记样本用于单类分类器训练。考虑到地表类型复杂，随机采样包含的类型有限，因此本研究采用多次分类并投票的原则，随机抽样 11 次获得 11 套未标记样本集,与果树/柑橘样本集进行分类,训练得到 11 次分类结果,

投票得到最终的分类结果。

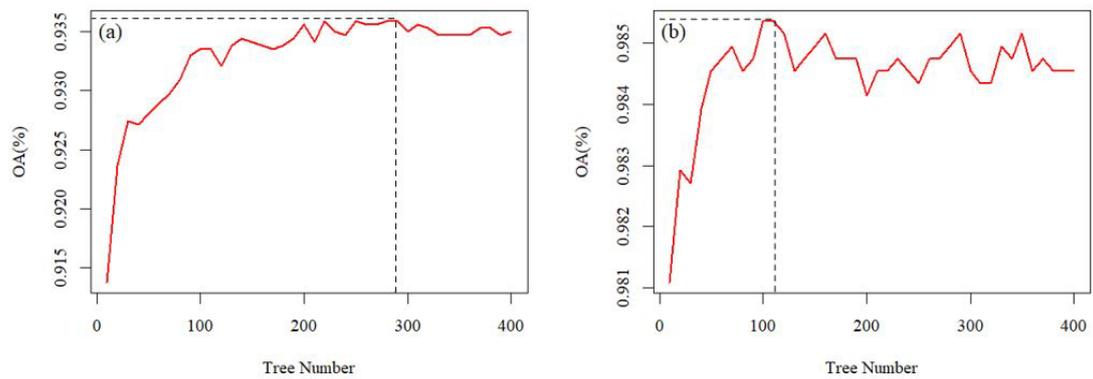

图 7 确定最优随机森林参数。 (a) 枝江分类器最优树选择图；(b) 桓台分类器最优树选择图;

然后，将训练好的随机森林分类器应用于研究区的作物识别。考虑到柑橘和果树分类器所得到的分类结果不确定性更小，本研究将识别出的柑橘制图结果，直接覆盖在湖北枝江作物的制图结果上。同理，将果树制图结果直接覆盖在山东桓台作物的制图结果上。结果如图9、图 10 所示。

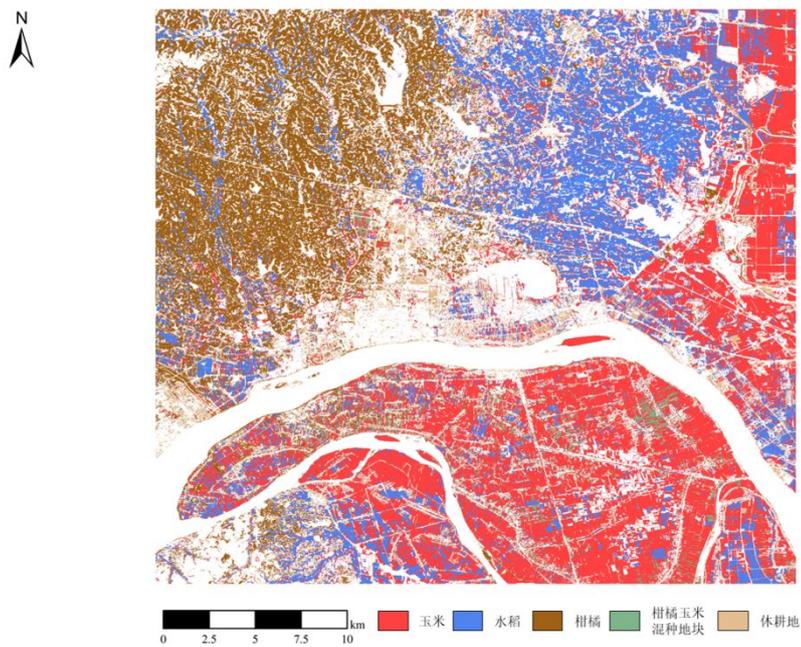

图 8 湖北枝江作物分类结果

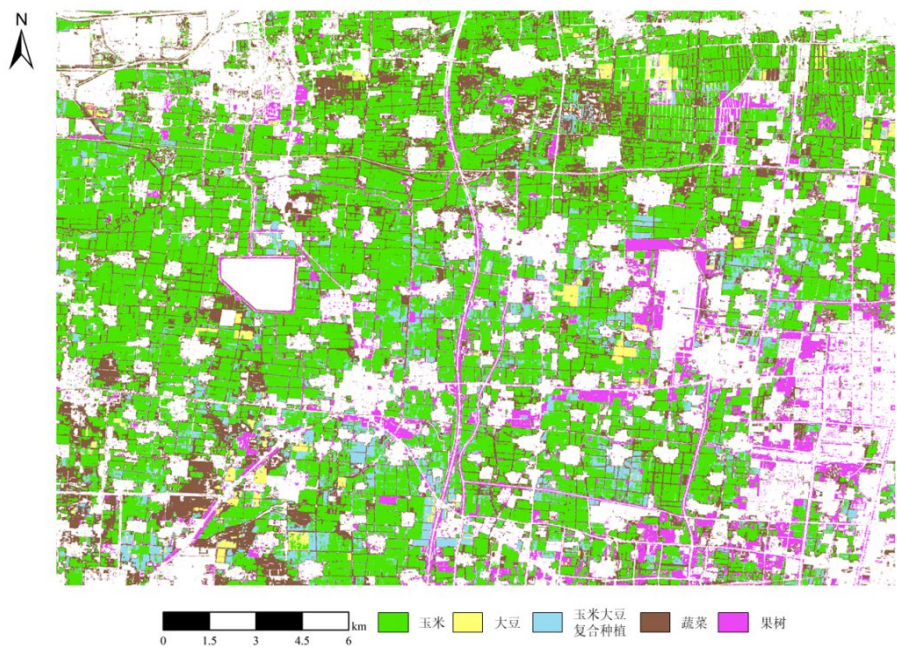

图 9 山东桓台作物分类结果

# 3.创新性

  本方法在协调运用"弱监督强特征"-"强监督弱特征"此级联分类框架的基础上，进一步优化领域知识的运用,对多种已有分类方法或数据产品取交的方式获得更为准确的强特征样本，并充分考虑云分布情况，生成更符合实际数据缺失分布的弱特征样本。从而进一步提升"弱监督强特征"-"强监督弱特征"级联分类框架的准确性与实用性。针对研究区云分布情况，随机去除一定数量"强特征"样本时序的有效观测，以生成符合实际数据缺失的"弱特征"样本。

# 4.场景通用性

  本研究提出的集成领域知识的"弱监督强特征"-"强监督弱特征"级联分类框架具有较强的场景通用性。第一，该分类框架具有较好的区域迁移能力和年际迁移能力。该框架在样本生成时弱化了特征，在一定程度上克服了由区域变化和时间变化导致的特征变化。本方法在MAP2021年作物识别大赛决赛中取得了82%的分类精度。第二，在"弱监督强特征"和"强监督弱特征"步骤中，分类器的选择灵活多样。"弱监督强特征"步骤基于小样本或专家知识，使用的分类器可以是自定义决策树,也可以选择对样本需求较少的其他机器学习算法如非监督学习、迁移学习、主动学习等方法（Wang et al., 2020）。而由于进行了特征弱化并生成了大量的样本，"弱特征强监督"步骤可以使用容量更大、学习能力更强的分类器而不必担心样

本数量不够造成的过拟合，例如常见的深度神经网络，包括 1D-CNN，RNN 和 LSTM 等 (Fawaz et. al )。第三，领域知识的运用减少了样本的区域依赖，提高了不同地区分类制图的可行性。

## 5.技术经济性

本研究提出的"弱监督强特征"-"强监督弱特征"级联分类框架十分经济便捷，其优势在于：一、通过领域知识的集成运用，克服足量样本的获取难题；二、我们采用的数据和整个算法均基于 Google Earth Engine 平台，均可实现云计算方便快捷。三、我们的模型参数容易获取，且容错性强。四、如 4 中所述，我们的模型有一定的区域迁移能力和年际迁移能力，因而可以用该模型解决更多的问题。

## 6.应用价值

本算法框架具有良好的创新性、场景通用性及技术经济性，可以较小的人力物力成本完成作物制图的任务。此外，用户可根据自己的时间及其它成本预算灵活的在本算法框架内选择不同的策略解决自己的问题，因而本方法的应用价值较高。

## 7.附录

本研究在"强特征弱监督"分类阶段决定利用专家知识、时序数据、谷歌影像，通过设定简单严格规则完成该阶段的作物像元提取。规则的设定主要源自于已有文献调研与之前研究的相关经验：

（1）收集专家知识：了解现有的文献来获取各个作物常用的特征；收集湖北省宜昌市和山东淄博市各个作物的农时，来大致把握各个作物的物候。

（2）寻找简单有效的区分特征：分析各个作物的物候历与作物相关特点，寻找能够区分玉米、大豆、水稻、柑橘、休耕地、玉米、蔬菜、果树等作物的遥感指数与时序特征。

（3）减少错分容许漏分：由于该阶段提取结果用于下一阶段分类的样本，因此严格设定规则，以保证不错分成其他作物类型或其他地表类型。同时严格设定规则实际上减少了经验决策树分类中的准确阈值设定的困难。

一、收集专家知识

本次比赛的识别作物为玉米、大豆、水稻、柑橘、休耕地、蔬菜、果树以及玉米大豆复种和柑橘玉米混种，经过文献调查得知：针对玉米、水稻和大豆的识别，不少研究表明，红边波段和短波红外波段是区分玉米、大豆和其它作物的重要特征，水稻在淹水期表现出独有的光谱特征，这是与其他作物区分的重要特征。You et al.（2020）利用玉米在收获前 60 天的 RENDVI 高于水稻和大豆，大豆的 RE2 波段在结荚期前后形成显著的峰值，水稻在淹水

期的 LSWI 显著高于其他作物，这为玉米、水稻和大豆找到了不同的最早可识别时间和可用特征。针对柑橘、蔬菜、果树、休耕地和混种，目前的相关遥感识别文献较少，方法仍待发展。

在物候信息收集上，我们通过中华人民共和国农业农村部的信息日历（http://zdscxx.moa.gov.cn:8080/nyb/pc/calendar.jsp）以及搜索中国种植信息，获取了湖北省 2022 年玉米、水稻及柑橘的详细农时和山东省 2022 年玉米、大豆、蔬菜、果树的详细农时。整理得到湖北省 2022 年玉米、水稻及柑橘的物候信息（图 A1）。桓台玉米为夏玉米，在六月种植，九月成熟；水稻在四月播种，从八月中旬到九月份开始抽穗、灌浆、成熟；柑橘全年都在生长，不同品种在全年不同季节成熟。经过整理得到山东省 2022 年玉米、大豆、蔬菜、果树的物候信息（图 A2）。春玉米是在四月底五月初播种，九月到十月成熟；夏玉米在六月份播种、出苗，九月成熟；大豆在五月份播种，九月下旬成熟；芹菜一年有两季，春季芹菜在二到三月份播种，六到七月份收获，秋季芹菜在七到八月份播种，冬季收获；葱在五月播种，九月到十月收获；梨树在春季发芽，夏季成熟，秋季落叶进入休眠期。

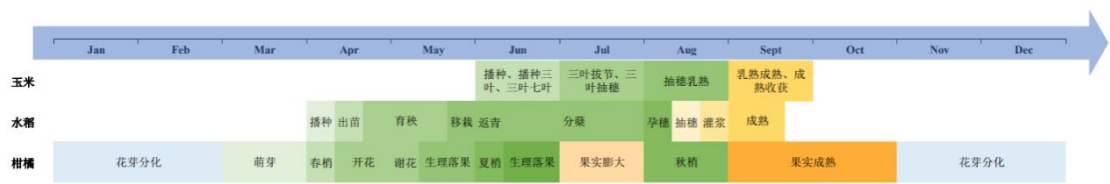

图 A1 湖北枝江 2022 年物候历

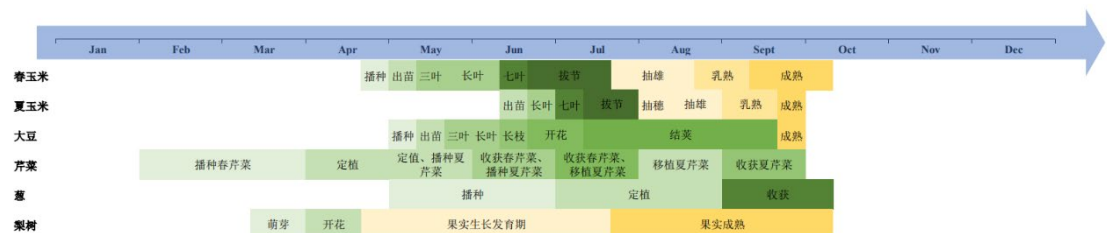

图 A2 山东桓台 2022 年物候历

## 二、确定分类特征

根据图 A2 可知，水稻在 4 月份播种，在四月下旬到五月份存在淹水期，这是与其他作物进行区分的最佳时段，此时段内水稻的 NDVI 低于其他作物，也低于 LSWI，利用此特征可以把水稻分离出来。针对夏玉米，我们引入红边波段，将此作为区分玉米和其他作物的重要特征。针对休耕地，根据休耕地的 NDPI 存在远远低于其他作物的特点进行区分。由 A3 和 A5 图可知，桓台县大豆在 6 月下旬至 7 月上旬播种，9 月上旬收获，在结荚期前后的 RENDVI 较低，以此区分大豆和玉米。结合物候信息我们发现桓台春玉米的 NDVI 曲线具有显著的特异性，根据玉米播种早、周期长、持续增长特点和蔬菜生长周期短特点，区分桓台区域内的玉米和蔬菜。针对湖北枝江柑橘和山东桓台果树，我们下载对应区域的谷歌影像，利用卷积神经网络进行识别。

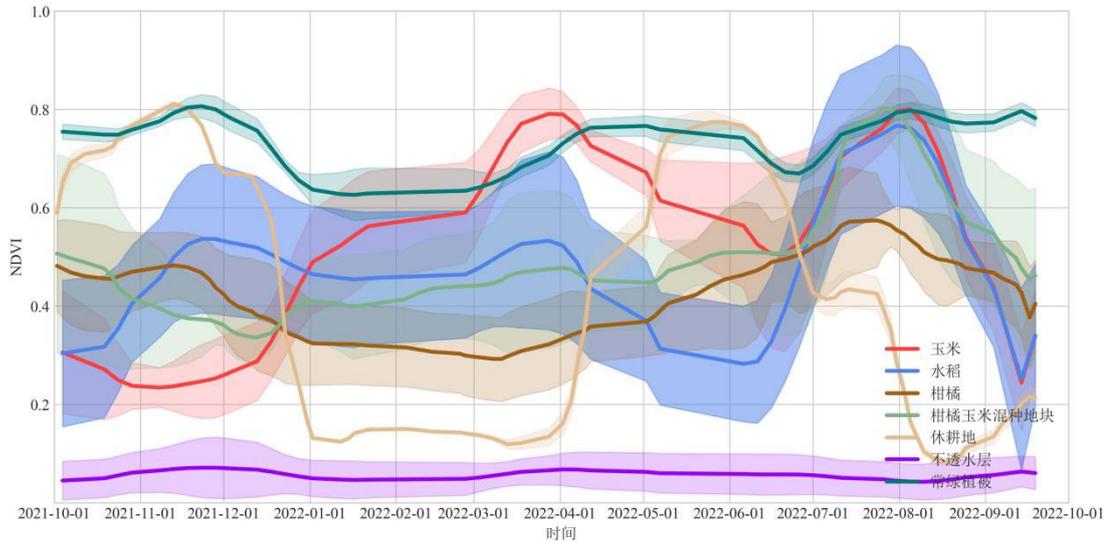

图 A3 基于 2022 年湖北少量样本的时间序列图

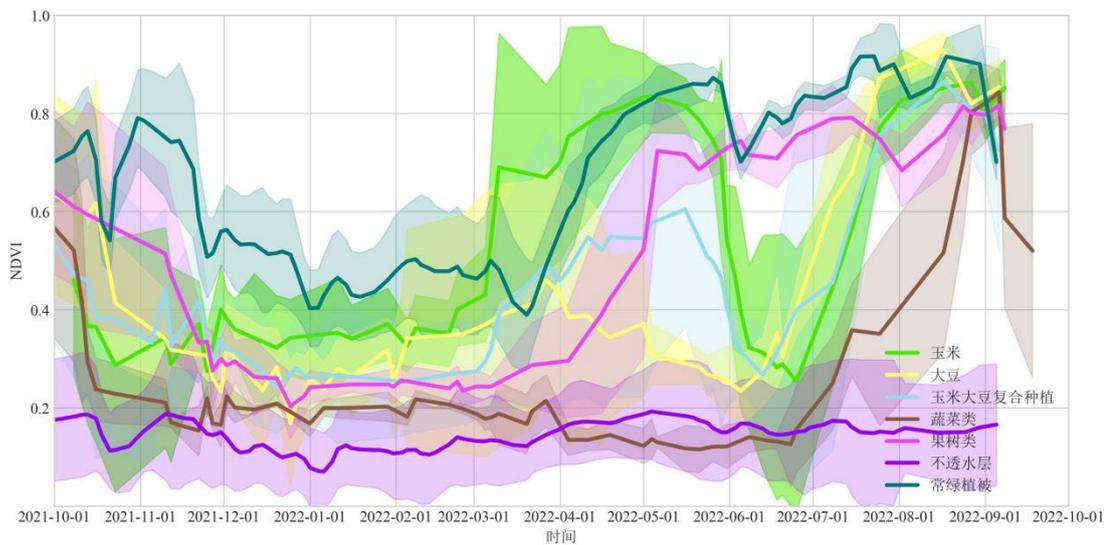

图 A4 基于 2022 年山东少量样本的时间序列图

### 三、决策树规则设定

#### 1. 休耕地提取规则

休耕地是耕地休耕制度下产生的一种耕地在可种作物的季节只耕不种或不耕不种的方式，其目的主要是使耕地得到休养生息，以减少水分、养分的消耗，并积蓄雨水，消灭杂草，促进土壤潜在养分转化，为以后作物生长创造良好的土壤环境和条件（杨庆媛等，2018）。休耕地是有计划、有规律地对肥力较差的耕地暂时性的退出粮食种植，待地力恢复后再重新耕耘的土地，其主要特点是在休耕期存在裸露的土壤，在植物的生长季节存在着较为稀疏的植物覆盖（Tong et al，2018）。其次，从时间的角度来看，休耕地与作物田具有不同的季节特征，这将是区分休耕地与有作物耕地的重要条件（Tong et al，2017）。同时，休耕地又与撂荒地不同，撂荒地存在长期无作物种植而导致的土壤肥力无法保障的情况，因此荒地长期不被作物覆盖的特点就成为了区分休耕地和荒地的重要条件。因此将休耕地与其他地物进行

区分存在以下挑战：① 要在一年中的某个时间段获取影像，以最佳方式捕捉休耕地与作物田的季节性差异；② 根据荒地长期不被作物覆盖的特点，从过去一整年的影像中捕捉休耕地和荒地的植物生长周期差异。

在研究中我们发现休耕地的 NDVI 和 NDPI 存在远远低于其他作物，但仅利用 8 月的 NDVI 和 NDPI 无法将休耕地与刚刚播种的作物区分，例如 7 月份刚播种的作物，其 8 月份的 NDVI 和 NDPI 值同样很低，因此需要将观测时段延长至 6 到 9 月，限制 NDPI 最大值小于 0.25。荒地常年没有作物生长，因此其 6 到 9 月 NDPI 值同样很低，可能与休耕地混分。我们认为休耕地与荒地的区别在于休耕地其他生长季会有作物生长，通过前一年有无作物生长（NDVI 高值持续超过一段时间）来排除休耕地。最终的决策规则如下

$$max\ (NDPI_t)_{t\in[150,290]} < 0.25 \quad (A\text{-}1)$$

$$(T_{NDVImax_{t\in[-365,0]}} - T_{NDVImin_{t\in[-365,0]}}) > 45 \quad (A\text{-}2)$$

**2. 玉米提取规则**

本次比赛要求提取 8 月份枝江地区和桓台地区种植的玉米。根据作物调查的结果和物候历，枝江地区种植夏玉米，在 5 月下旬至 6 月上旬播种，9 月中旬收获；桓台地区种植春玉米、夏玉米和大豆。其中，春玉米在 4 月播种，8 月收获；夏玉米在 6 月下旬至 7 月上旬播种，9 月下旬收获。

依据玉米特定的物候特征，不少玉米提取方法被提出。Qiu 等人依据玉米在抽穗期前后 NMDI（Normalized Multi-band Drought Index）和 EVI2 特定的曲线模式，构造了 RCPN（Cumulative Positive slope to Negative）指数用于提取全国的玉米。Zhong 等人依据玉米和大豆的物候特征，制定了一系列样本选取规则，自动选取玉米和大豆样本输入最大似然分类器进行分类。王雪婷等人使用曲线匹配的方法提取了黄淮海区域夏玉米的种植面积。然而，上述方法均需要较完整的玉米物候曲线，目前本次比赛的研究区内夏玉米还未收割，因此，这些方法不适用于 2022 年 8 月份的夏玉米的提取。

幸运的是，不少研究表明，红边波段和短波红外波段是区分玉米和其它作物的重要特征。图 A5 是 You 等人绘制的三江平原玉米、水稻和大豆的时序曲线。研究表明，在 DOY 约为 210 天前后时，玉米的 RENDVI 高于水稻和大豆；DOY 在 120-150 天时，水稻的 LSWI 显著高于玉米和大豆的 LSWI。基于以上先验知识和枝江地区的物候历（图 A1），我们设定了枝江的夏玉米提取规则：

$$max\ RENDVI_{t\in[180,240]} > 0.18 \quad (A\text{-}3)$$

$$min\ (NDVI_t)_{t\in[120,150]} < 0.4 \quad (A\text{-}4)$$

$$max\ (NDVI_t)_{t\in[150,240]} > 0.6 \quad (A\text{-}5)$$

$$LSWI_t < NDVI_t, t\in[90,150] \quad (A\text{-}6)$$

其中，公式 A-4 为了依据玉米抽穗期 RENDVI 高的特点，选取具有显著玉米特征的样本；公式 A-5 保证了夏玉米的播种时间，即播种时间为 6 月下旬-7 月上旬，同时由于果树此时 NDVI 较高，该公式也可以去除果树样本；公式 A-6 保证大豆在生长期的 NDVI 较高；

公式 A-7 依据水稻淹水期 LSWI>NDVI 的特点,去除水稻样本的干扰。对于提取出的像素,应用形态学开运算去除孤立的像素点。

桓台的夏玉米提取规则与枝江类似,各项规则的阈值保持不变,而规则应用的时间依据当地的物候历做出相应的调整:

$$max\ RENDVI_{t\in[210,270]} > 0.18 \quad (A\text{-}8)$$

$$min\ (NDVI_t)_{t\in[150,180]} < 0.4 \quad (A\text{-}9)$$

$$max\ (NDVI_t)_{t\in[180,270]} > 0.6 \quad (A\text{-}10)$$

$$LSWI_t < NDVI_t\ _{t\in[90,150]} \quad (A\text{-}11)$$

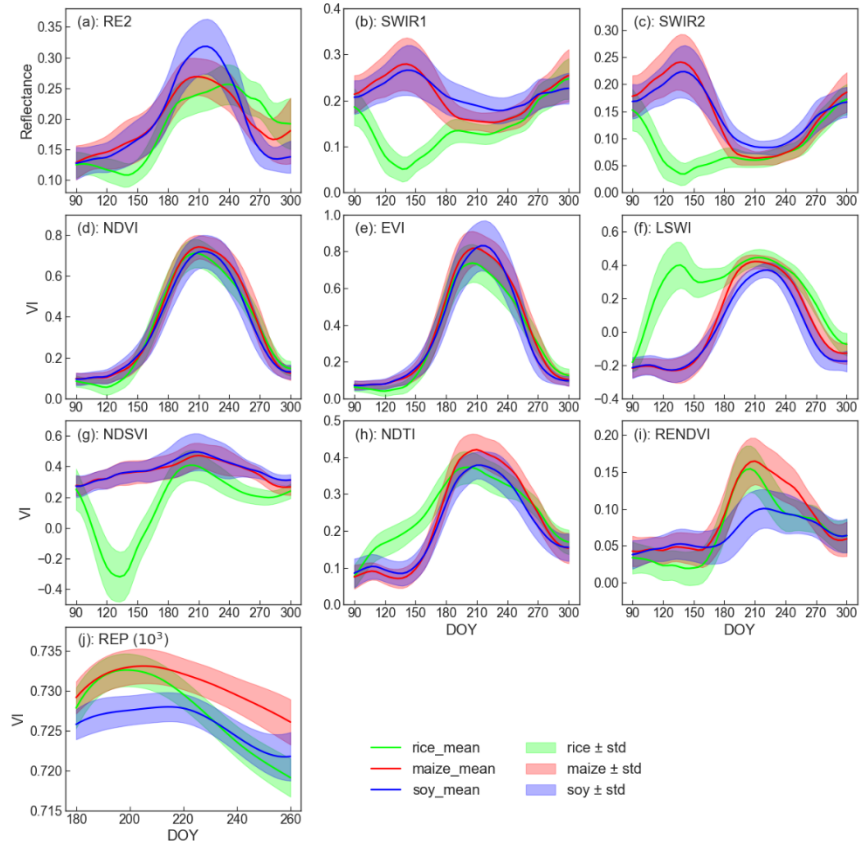

图 A5 三江平原玉米、水稻和大豆的时序曲线

桓台 8 月份同时种植春玉米,因此,我们使用初赛的玉米提取规则提取春玉米。春玉米具有播种早、周期长和 NDVI 持续增长的特点,基于此我们制定了以下春玉米提取规则。

$$max\ (NDVI_t)_{t\in[60,120]} < 0.25 \quad (A\text{-}12)$$

$$max\ (NDVI_t)_{t\in[90,150]} > 0.3 \quad (A\text{-}13)$$

$$max\ (NDVI_t)_{t\in[150,210]} > 0.5 \quad (A\text{-}14)$$

$$DOY_{max} - DOY_{min} \geq 100 \quad (A\text{-}15)$$

式 A-12,A-13 和 A-14 基于春玉米播种早和持续增长的特点,要求作物在 3-4 月的 NDVI 最大值小于 0.25,4-5 月的 NDVI 最大值大于 0.3,6-7 月的 NDVI 最大值大于 0.5。由于春玉米的生长期长度长于蔬菜,因此 A-15 应用生长期长度(即 6-7 月 NDVI 最大值的

时间$DOY_{max}$和3-4月NDVI最小值时间$DOY_{min}$之差）不小于100天区分春玉米和蔬菜。

### 3. 大豆提取规则

本次比赛要去提取8月份桓台县种植的大豆。依据物候历，桓台县大豆在6月下旬至7月上旬播种，9月上旬收获。依据You等人的研究，三江平原大豆的RE2波段在DOY为210天前后形成显著的峰值，而水稻和玉米值都较低。此外，在该日期前后，大豆的RENDVI较水稻和玉米低。依据这些特点和桓台地区的物候历，我们制定了大豆样本的提取规则。

$$max\ RE2_{t\in[210,240]} > 0.45 \quad (A\text{-}16)$$

$$max\ (RENDVI_t)_{t\in[210,240]} < 0.15 \quad (A\text{-}17)$$

$$min\ (NDVI_t)_{t\in[150,180]} < 0.4 \quad (A\text{-}18)$$

$$max\ (NDVI_t)_{t\in[180,270]} > 0.6 \quad (A\text{-}19)$$

其中，RE2为Senitnel-2 B6波段，公式A-16用于选择B6波段具有高反射率的大豆样本；由于玉米的RENDVI较高，而大豆的RENDVI较低，公式A-17基于此去除异常的玉米样本；A-18和A-19与夏玉米同理，分别用于保证大豆在6-7月种植和大豆的生长及NDVI较高。对于提取出的像素，应用形态学开运算去除孤立的像素点。

### 4. 水稻提取规则

水稻的物理特性是稻株在淹水土壤中生长，特别是当水稻田处于淹水期和插秧期时，水稻田的地表为地表水与绿色水稻的混合状态。而这一特点是遥感提取水稻最为显著的特征，显示出了独特的光谱特征、微波散射特征等（Xiao et al, 2005；Pei et al, 2021）。在现有的遥感提取水稻的研究中（Zhang et al, 2015; Liu et al, 2018），由于LSWI的时间动态变化可以用来捕捉水稻淹水和移栽引起的地表水增加，因此学者们常利用LSWI与植被指数（NDVI、EVI）相结合来识别水稻，并取得良好的提取效果。例如，Xiao等人（2005）设定淹水期与插秧期中符合LSWI+0.05≥NDVI为水稻，邓刚等人（2020）直接利用LSWI与EVI的时间序列识别水稻。湖北省2022年的水稻在4月中下旬至5月下旬进行育秧和移栽，因此制定规则如下：

$$count\ (LSWI_t + 0.05 \geq NDVI_t)_{t\in[105,150]} > 2 \quad (A\text{-}20)$$

$$max\ (NDVI_t)_{t\in[0,290]} > 0.4 \quad (A\text{-}21)$$

$$min\ (NDVI_t)_{t\in[0,290]} > 0.05 \quad (A\text{-}22)$$

### 5. 蔬菜提取规则

目前专门针对芹菜、大葱遥感提取的文献较少，方法仍待发展，我们结合桓台县实地调研和蔬菜有别于粮食作物生长特点设定提取规则。通过实地调研可知，桓台县下半年的芹菜和大葱大部分是五六月种植，大葱播种后秋季约80-95天收获，芹菜播种后约100天后收获，由于芹菜和大葱物候期相似，采取相同的提取规则。

蔬菜发芽到幼苗期大约 40 天，一般播种后（6-7 月）蔬菜 NDVI 值较低，因此可以与果树区分。同时蔬菜出苗生长快，6-7 月 NDVI 最大值大于 0.5。

蔬菜提取的关键规则是雷达特征和生长旺盛期提取。雷达特征选择微波极化差异指数 MPDI (Microwave Polarization Difference Index)，该指数对植物含水量和土壤湿度的变化相当敏感(Chauhan and Srivastava, 2016)。由于蔬菜在出苗期灌溉充分，土地接近裸土，而玉米、大豆出苗期对水分需求相对于蔬菜较低，与蔬菜冠层结构也不同，在出苗期至生长旺盛期玉米和大豆 MPDI 表现出上升趋势，而蔬菜表现出下降趋势，从而提取出蔬菜。尽管对雷达数据进行了预处理和滤波去噪，但是由于 Sentinel-1B 合成孔径雷达成像卫星在 2021 年 12 月发生故障，2022 年的 Sentinel-1 雷达数据存在一定波动和缺失，不能完全提取出蔬菜，还需要其他重要特征进行补充。

因此参考结合物候的大宗作物和小宗作物同时提取(Goldberg et al., 2021)，我们进行了蔬菜生长旺盛期的提取。蔬菜在 7 月中旬以后开始进入生长旺盛期，设置蔬菜八月 NDVI 最大值大于 0.6，且最小值大于 0.2，确保蔬菜仍在生长没有被收割。对于蔬菜生长旺盛期提取，设定 1 月 1 日-9 月 10 日蔬菜 NDVI75%分位数为生长旺盛期开始，6 月 1 日-9 月 10 日 NDVI 最大值日期为生长旺盛期结束，蔬菜旺盛期长度小于 45 天，而玉米、大豆以及玉米大豆混种的生长旺盛期比蔬菜更长，可以很好提取出蔬菜。

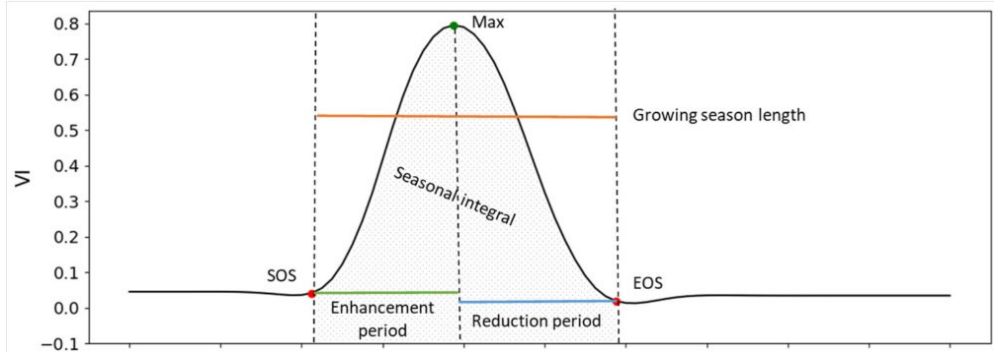

图 A6 一般作物的物候特征(Goldberg et al., 2021)

$$\max(NDVI_t)_{t\in[151,211]} > 0.5 \tag{A-22}$$

$$MPDI = \frac{\sigma^\circ_{VV} - \sigma^\circ_{VH}}{\sigma^\circ_{VV} + \sigma^\circ_{VH}} \tag{A-23}$$

$$15\%\min(NDVI_t)_{t\in[151,211]} < 0.25 \tag{A-24}$$

$$MPDI_{max_{t\in[134,165]}} - MPDI_{min_{t\in[195,211]}} > 0 \tag{A-25}$$

$$NDVI_{max_{t\in[212,242]}} > 0.6 \;\&\&\; NDVI_{min_{t\in[212,242]}} > 0.2 \tag{A-26}$$

$$length(T_{NDVImax_{t\in[151,252]}} - T_{NDVIp75_{t\in[1,252]}}) < 45 \tag{A-27}$$

**6. 果树提取规则**

本文基于深度神经网络提取果树（柑橘和梨树）样本，所使用的是经典卷积神经网络 VGGNet19。VGG 是牛津大学计算机视觉组（Visual Geometry Group）提出的深度卷积神经

网络(Simonyan K et al, 2014)。该模型参加 2014 年 ImageNet 图像分类与定位挑战赛，在分类任务上排名第二，在定位任务上排名第一。同年 GoogleNet 获得了分类任务的第一名，但在多个迁移学习任务中 VGG 表现更好。VGG 中根据卷积核大小和卷积层数目的不同，可分为 A，A-LRN，B，C，D，E 共 6 个配置(ConvNet Configuration)，其中以 D，E 两种配置较为常用，分别称为 VGG16 和 VGG19（如图 A7）。VGG19 包含了 19 个隐藏层（16 个卷积层和 3 个全连接层），相比 VGG16 增加了 3 个卷积层（如图 A8），层数更深，泛化性能更强。VGG 结构中采用了小卷积核和小池化核的策略，卷积核大小都是 3x3，池化核的大小都是 2x2，对于给定的感受野，使用堆叠的小卷积核是优于采用大的卷积核，多层非线性层的堆叠能够增加网络深度，从而保证模型能够学习更复杂的模式，而且因为小卷积核的参数更少，所以整个网络的计算代价较小。

| ConvNet Configuration | | | | | |
|---|---|---|---|---|---|
| A | A-LRN | B | C | D | E |
| 11 weight layers | 11 weight layers | 13 weight layers | 16 weight layers | 16 weight layers | 19 weight layers |
| input (224 × 224 RGB image) | | | | | |
| conv3-64 | conv3-64 LRN | conv3-64 conv3-64 | conv3-64 conv3-64 | conv3-64 conv3-64 | conv3-64 conv3-64 |
| maxpool | | | | | |
| conv3-128 | conv3-128 | conv3-128 conv3-128 | conv3-128 conv3-128 | conv3-128 conv3-128 | conv3-128 conv3-128 |
| maxpool | | | | | |
| conv3-256 conv3-256 | conv3-256 conv3-256 | conv3-256 conv3-256 | conv3-256 conv3-256 conv1-256 | conv3-256 conv3-256 conv3-256 | conv3-256 conv3-256 conv3-256 conv3-256 |
| maxpool | | | | | |
| conv3-512 conv3-512 | conv3-512 conv3-512 | conv3-512 conv3-512 | conv3-512 conv3-512 conv1-512 | conv3-512 conv3-512 conv3-512 | conv3-512 conv3-512 conv3-512 conv3-512 |
| maxpool | | | | | |
| conv3-512 conv3-512 | conv3-512 conv3-512 | conv3-512 conv3-512 | conv3-512 conv3-512 conv1-512 | conv3-512 conv3-512 conv3-512 | conv3-512 conv3-512 conv3-512 conv3-512 |
| maxpool | | | | | |
| FC-4096 | | | | | |
| FC-4096 | | | | | |
| FC-1000 | | | | | |
| soft-max | | | | | |

图 A7 VGG 网络结构

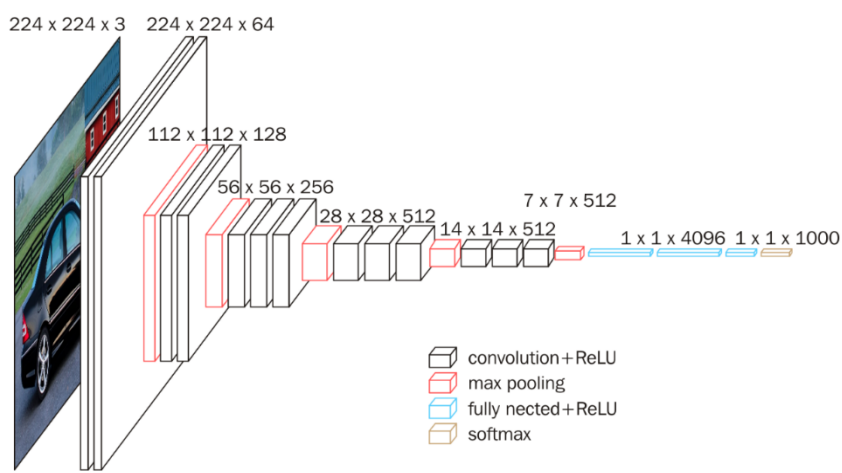

图 A8 VGG16 可视化网络结构

本文采用迁移学习的思想，即将训练好的模型（预训练模型）参数迁移到新的模型来优化新模型训练。首先我们使用 VGG19 在 ImageNet 上的预训练权重模型文件初始化模型参

数，接下来，使用准备好的待分类图像数据集训练和微调 VGG 模型的权重。最后将测试图像直接送入训练好的 VGG19 网络，然后会输出图像所属类别以及置信度。图像数据采用的是 GoogleEarth 空间分辨率为 0.6m 的影像，图像大小为 50m x 50m。目视解译选出少量影像作为训练数据集，用于训练和微调模型 VGG 模型的参数。待预测的样本图像依据 global30，对各类别进行分层随机抽样。柑橘图像的抽样范围是研究区范围，梨树图像的抽样范围是研究区及 10km 缓冲区。柑橘抽样生成 32000 张影像，梨树抽样生成 87000 张影像。使用训练好的 VGG 模型对影像进行预测，其中分类为柑橘/梨树的影像作为样本影像。分别对柑橘和梨树训练两个 VGG 模型，训练数据集共包含 8 类遥感影像，分别是：柑橘（115）/梨树（75）、云污染（185）、水体（60）、雪（131）、草地（130）、森林（400）、耕地（147）、城市（90），其中使用 70%的数据作为训练集，30%的数据作为验证集。本文在 pytorch 框架下训练 VGG19 模型，训练时的 batch size 设置为 16，epoch 设置为 100，初始学习率为 0.0001。训练结果显示柑橘验证集精度为 0.99，梨树验证集精度为 0.98。预测的影像中 VGG 模型预测出 3017 张柑橘影像，212 张果树影像。

预测出的果树的影像进行后处理，以提高样本的精确性及可靠性。使用 Grad-CAM(Gradient-weighted Class Activation Map)提取每个图像中的果树范围，细化果树区域，将各图像中提取的果树范围合并作为最终的果树样本。Grad-CAM 方法对输入图像生成类别激活的热力图。它是与特定输出类别相关的二维特征分数网络，网格的每个位置表示该类别的重要程度(Selvaraju R R et al, 2017)。对于一张输入到 CNN 模型且被分类成"果树"的图片，该技术可以以热力图形式呈现图片中每个位置对"果树"类别的特征贡献程度。有助于了解一张原始图像的哪一个局部位置让 CNN 模型做出了最终的分类决策（如图 A10 所示）。特征贡献度高的区域说明与"果树"的特征更为相似，该区域为"果树"的置信度更高。我们选择 VGG19 的最后一个卷积层的输出特征图，得到其对应的热力图。这样可以得到最终决定分类类别的关键特征区域。我们将 VGG 预测为果树的图像作为 Grad-CAM 的输入，获取各区域对果树类别的特征贡献程度，提取每张图像中对"果树"类别贡献度高的区域，以阈值 0.5 划分出每张图像中的果树区域，仅以果树区域作为样本。预测得到的果树影像并不是均质的果树图像，可能包含多种地物，其中仅部分区域为果树。我们用该技术能够更加严格的筛选样本，提取图像中果树类别置信度高的区域，可以更精细化果树区域样本，大大降低影像内混合地物来的影响，提高了样本区域的精度和可靠性。其核心公式为：

$$\alpha_k^c = \frac{1}{z}\overbrace{\sum_i \sum_j \frac{\partial y^c}{\partial A_{ij}^k}}^{\text{global average pooling}} \tag{A-29}$$

$$L_{Grad-CAM}^c = ReLU\underbrace{\left(\sum_k \alpha_k^c A^k\right)}_{\text{linear combination}} \tag{A-30}$$

$\frac{\partial y^c}{\partial A_{ij}^k}$ 为 c 类中第 k 个特征图的权重；c：第 c 类别，k：最后一个卷积层特征图通道。

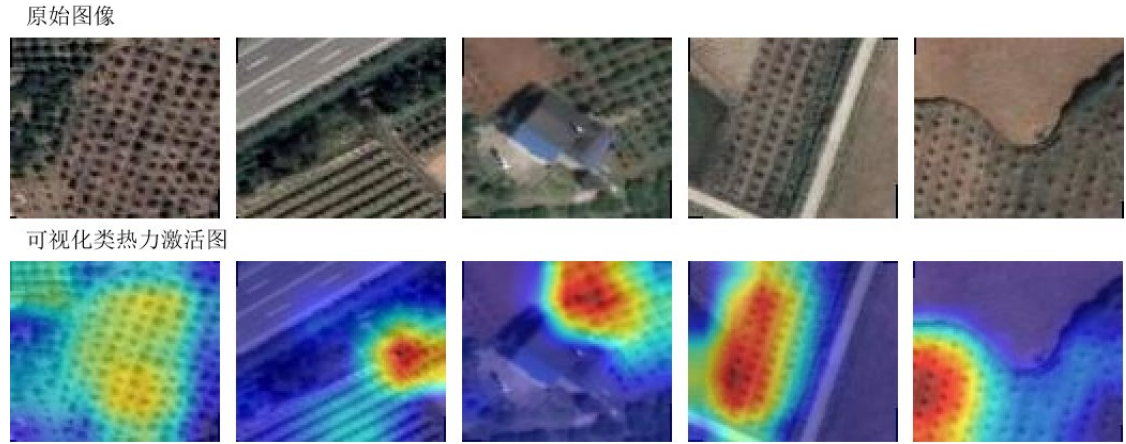

图 A9 热力图可视化特征贡献程度

**7. 混种提取规则**

玉米-大豆带状复合种植技术能够实现粮食增产和土地高效利用，在近五年得到广泛应用。2022 年，山东省首次大面积推广该技术。遥感以像元为最小观测单位，无人机影像能够清晰地拍摄出研究区的复合种植模式，但需花费时间和经济成本。本团队采用 Sentinel-2 高分影像具有 10m 的高分辨率，但在玉米大豆复合种植的情境下，不能获取纯净的玉米与大豆像元。因此，我们采用几何光学模型，通过对山东省 6-8 月的太阳高度角计算、复合种植模式估计、作物生长模式估计，生成考虑光照和阴影的玉米大豆混合像元样本。

- 太阳高度角

太阳高度角，即太阳光线与地平面的夹角，受日期、时间、经纬度等多因素影响。桓台市经纬度约为（118°E, 37°N），Sentinel-2 在桓台市的卫星过境时间约 11 点。计算 6 月 2 日-9 月 10 日的太阳高度角，计算结果如下表。

表 3 太阳高度角计算结果

| 日期 | 6-2 | 6-12 | 6-22 | 7-2 | 7-12 | 7-22 | 8-1 | 8-11 | 8-21 | 8-31 | 9-10 |
| --- | --- | --- | --- | --- | --- | --- | --- | --- | --- | --- | --- |
| DOY | 183 | 193 | 203 | 213 | 223 | 233 | 243 | 253 | 263 | 273 | 283 |
| 太阳高度角/° | 69.78 | 68.97 | 67.64 | 65.81 | 63.49 | 60.75 | 57.66 | 54.30 | 50.75 | 47.11 | 43.48 |

- 复合种植模式

复合种植模式即大豆与玉米的比例，主要包括"3:2 模式"、"4:2 模式"、"4:3 模式"、"6:3 模式"等，在山东省多为 2 行玉米搭配 2-4 行大豆。本团队综合官方农业网站、新闻、论文等多渠道的信息，采用 3:2 的模式开展模拟。

- 作物生长模式

综合研究区实况与专家经验，采用分段线性模型模拟玉米和大豆的作物生长模式。随日期变化，玉米的作物高度$h_{maize}$和大豆的作物高度$h_{soybean}$模拟如下（单位：m）。

$$h_{maize} = \begin{cases} 0.08 * (T - 183), \ 183 \leq x < 223 \\ 0.12 * (T - 223) + 0.64, \ 223 \leq x \leq 283 \end{cases} \quad \text{(A-31)}$$

$$h_{soybean} = \begin{cases} 0.04 * (T - 183), & 183 \leq x < 223 \\ 0.12 * (T - 223) + 0.64, & 223 \leq x \leq 283 \end{cases} \quad \text{(A-32)}$$

- 考虑阴影的混合像元模型

由于玉米的生长速度更快、最终生长高度更高,需考虑玉米阴影对大豆作物的遮挡效应。图 A10 给出了两种情景下的玉米-大豆带状复合种植的几何光学模型。6 月初,玉米和大豆的高度较矮,而太阳高度角较高,此时玉米的阴影较短,不能对大豆顶部形成遮挡。8 月时,玉米和大豆的高度较高,且玉米较大豆更高,而太阳高度角较低,此时玉米的阴影高且长,对大豆顶部可能形成遮挡。

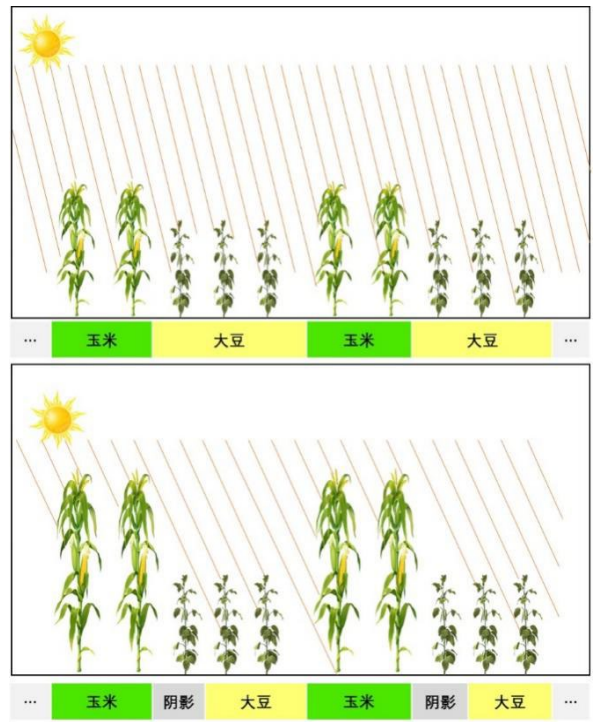

图 A10 玉米-大豆带状复合种植的几何光学模型

阴影遮挡的判断规则见图 A11,图中Δx为田垄宽度(估算为 0.6m),α 为太阳高度角(计算结果见表 1)。当阴影高度高于大豆高度时($h_{shade} > h_{soybean}$),该株大豆被遮挡,在混合像元中表现为阴影区域。6 月和 7 月,由于玉米和大豆无遮挡效应,此时的玉米和大豆在混合像元中占比分别为 40%和 60%。自 8 月 21 日起,形成的混合像元中玉米和大豆占比变为 40%和 40%,另外 20%为阴影遮挡区域。自 9 月 10 日起,形成的混合像元中玉米和大豆占比变为 40%和 20%,另外 40%为阴影遮挡区域。

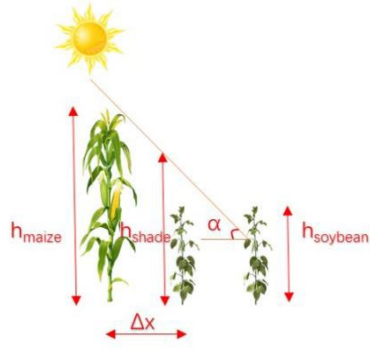

图 A11 阴影遮挡的判断规则

最后，采用线性混合像元模型，逐日期生成玉米大豆复合种植样本。混合像元模型公式如下，式中，S 表示混合像元反射光谱，$f_i$ 和 $x_i$ 分别表示第 i 个端元的端元丰度和光谱，w 表示误差。在本应用场景中，取玉米、大豆、阴影三类端元。

$$S = \sum_{i=1}^{m} f_i x_i + w \tag{A-33}$$

$$\sum_{i=1}^{m} f_i = 1 , f_i > 0 , i = 1, \dots, n \tag{A-34}$$

与玉米大豆复合种植的提取类似，对柑橘玉米混种的提取，也采用混合像元模型进行模拟。首先构建了几何光学模型，计算湖北省 6-8 月的太阳高度角，并确定了玉米柑橘"1:1"的混种模式。然后，考虑到玉米和柑橘树的实际高度，经模拟，认为玉米和柑橘无阴影遮挡效应，即按照 50%和 50%的比例分别估算玉米和柑橘。最后，采用线性混合像元模型，生成了柑橘玉米混种样本。

# 参考文献